\documentclass{llncs}

\usepackage{epsfig} %% for loading postscript figures
\usepackage{enumitem}
\usepackage{amsmath}
\usepackage{graphicx}
\usepackage{subfigure}
\usepackage{diagbox}
\usepackage{makecell}
\usepackage{cite}
\usepackage{array}

\title{
Sketch-Inspector: a Deep Mixture Model for High-Quality Sketch Generation of Cats}

\author{Yunkui Pang \inst{1} \and Zhiqing Pan \inst{2} \and Ruiyang Sun \inst{3} \and Shuchong Wang \inst{4}}

\institute{University of Southern California, Los Angeles, CA, USA \email{yunkuipa@usc.edu}
\and Sun Yat-sen University, Guangzhou, China
\email{panzhq@mail2.sysu.edu.cn}
\and University of Waterloo, Waterloo, Canada
\email{sunruiyang@live.cn}
\and Michigan State University, East Lansing, MI, USA \email{wangsh91@msu.edu}}

\begin{document}

\maketitle    

\begin{abstract}
{With the involvement of artificial intelligence (AI), sketches can be automatically generated under certain topics. Even though breakthroughs have been made in previous studies in this area, a relatively high proportion of the generated figures are too abstract to recognize, which illustrates that AIs fail to learn the general pattern of the target object when drawing. This paper posits that supervising the process of stroke generation can lead to a more accurate sketch interpretation. Based on that, a sketch generating system with an assistant convolutional neural network (CNN) predictor to suggest the shape of the next stroke is presented in this paper. In addition, a CNN-based discriminator is introduced to judge the recognizability of the end product. Since the base-line model is ineffective at generating multi-class sketches, we restrict the model to produce one category. Because the image of a cat is easy to identify, we consider cat sketches selected from the {\it QuickDraw} data set. This paper compares the proposed model with the original Sketch-RNN on 75K human-drawn cat sketches. The result indicates that our model produces sketches with higher quality than human's sketches.
}
\end{abstract}

\section{Introduction}
\noindent
Sketches are images drawn by humans with symbols, lines, strokes or shapes. Unlike real-life image, they do not always present the entire appearance of things, but can indicate not only the idea of artists but also the theme \cite{c_12}. Therefore, with the popularity of image generation, sketches, as a unique form of an image, receives people’s attention for its  abstractness.

Traditional generative models such as generative adversarial network \cite{c_7}, variational inference \cite{c_13}, and auto regressive models \cite{c_14} produce an image from the pixel level. Indeed, these models are capable of generating sketch-like images. However, they lose the essence of sketch by depicting objects by ordered strokes \cite{c_1}. Therefore, models based on sequential vectors instead of pixel image were developed for AI sketching.

Several breakthroughs have been made in AI sketching. Sketch-RNN \cite{c_1} was firstly proposed in 2017 to draw sketches like humans. Sketch-pix2seq \cite{c_2} in late 2017 added CNN layers to enhance model's generation ability over multiple categories. Jaques et al. \cite{c_5} used facial expression feedback to improve the model's accuracy in 2018. In the same year, Sketch-aae introduced Reinforcement Learning into sketch generation and trained a GAN model to sketch. Inspired by their insights, we have focused on one aspect that was ignored in previous research: let AI 'see' and evaluate what it has drawn and then adjust its previous behavior based on evaluation might improve the quality of generated sketches.

Our paradigm is based on the observation of the process of human sketching. We noticed that humans use erasers to erase what they have drawn and change the trail of stroke based on their general evaluation of the current drawing. This strategy gives human more opportunities to revise the painting, which gives them a greater likelihood to produce high-quality paintings. We want to examine the effectiveness of this strategy on AI sketching. Therefore, we proposed a CNN-based decoder to assist the variational autoencoder (VAE) based on recurrent neural network (RNN).

This paper makes the following contributions:

\begin{enumerate}
[topsep=0pt,itemsep=1pt,partopsep=0pt,leftmargin=1cm]
\item Proposed CNN network to enhance the quality of sketches produced by Sketch-RNN. The process adjusts the position of each stroke in the generated sketch.
\item Validated the performance of our system through multiple methods, including Discriminator model and t-SNE. Our model successfully misleads the discriminator to view generated sketches as the original one drawn by a human, and the rate is 46\% more than the original Sketch-RNN. The t-SNE result suggests that features of sketches produced by our model are more stable than those produced by Sketch-RNN.
\end{enumerate}
    
According to the original paper of the base-line model \cite{c_1}, one main drawback of the model is its low productivity to generate identifiable multi-class sketches. To guarantee accuracy, we use a pre-trained model that concentrates on one category. Considering that cats are relatively easier to recognize compared with complicated figures such as mermaids or lobsters \cite{c_1}, we choose cat sketches selected from {\it QuickDraw} data set.

%%%%%%%%%%%%%%%%%%%%%%%%%%%%%%%%%%%%%%%%%%%%%%%%%%%%%%%%%%%%%%%%%%%%%%
\section{Related Works}
\noindent
Ha et al.\cite{c_1} investigated Sketch-RNN to generate sketches similar to human-drawn sketches in 2017. The model is a sequence-to-sequence variational autoencoder. A novel representation of sketches, sequence of pixel coordinates of strokes, was proposed to satisfy the input format of the model. After that, several works managed to improve the method. Chen et al.\cite{c_2} substituted the bi-directional RNN encoder with a CNN encoder to predict mean and standard deviation of the Normal-distributed position of strokes in 2017. This helped the model to remember more sketches of different models while producing less recognizable sketches as a compromise. Wu et al. \cite{c_3} used Sketchsegenet to instruct RNN to sketch more like human, which can assist AI to understand the order of strokes and the meaning of each stroke. In 2018, Diaz Aviles and his lab \cite{c_15} developed a CNN-based GAN that generates sheep sketches. Cao et al. \cite{c_4} introduced a CNN-based method into the original Sketch-RNN, which applied a CNN model to help RNN-based encoder extract latent features from original sketches. The added CNN helped the model to remember more sketches from different categories and generate more recognizable sketches.

Other researchers tried different training processes to improve the model. Jaques et al.\cite{c_5} used facial expression from volunteers when they saw a sketch as a feedback to improve the model. They labeled the facial expressions into positive and negative ones. The model was designed to achieve positive feedback as much as possible. Another work \cite{c_6} used Reinforcement Learning accompanied with a GAN model to draw sketches. They proposed VASkeGAN based on Sketch-RNN. The model added a GRU based recurrent neural network as a discriminator. Policy gradient was used in training model and stroke proposal.

All these works explored the potential improvement directions with different models and achieved some amount of success. However, we found that no one considered an external feedback when the model is sketching. In order to achieve this goal, we added a CNN on the decoder part. Our work differs from the previous research by a simultaneous adjustment of strokes on generation from the aspect of reducing disordered strokes produced by the model.

\section{Methodology}

\subsection{Data set}
\noindent
We used the data set of cat sketches provided by {\it QuickDraw} from Google. It contains 100,000 of human drawn cat sketches. The format of the data set is 
\begin{math}
[\Delta x, \Delta y, p_1, p_2, p_3]
\end{math}. 
\begin{math}\Delta x \end{math} 
and 
\begin{math}\Delta y \end{math} 
denotes the distance from the previous point in the x-y coordinate. p$_{1}$, p$_{2}$ and p$_{3}$ represents the state of drawing process. p$_{1}$ means the current stroke continues to this point. p$_{2}$ means the end of the current stroke. p$_{3}$ means the end of the whole sketch.

The data set was divided into three parts: 70,000 training set, 2,500 test set and 2,500 validation set. 15,000 sketches from the training set were further randomly selected to train our model.

\subsection{Model}
\subsubsection{Sketch-RNN}
\noindent
Sketch-RNN is based on the variational autoencoder. The encoder of the model is a bidirectional RNN. It accepts a sequence of strokes in a format of 
\begin{math}
[\Delta x, \Delta y, p_1, p_2, p_3]
\end{math} 
as mentioned above. The output is also a sequence with the same format. Both of the sequences are organized in a time order.

To generate a sketch, the encoder first produces a hidden vector given sequences of strokes. The vector is the last output node in the encoder. Since the VAE model assumes that data follows normal distribution, the vector is further compressed into mean and standard deviation of the distribution. Then random sampling under this normal distribution is used to produce the input of the decoder. The decoder is composed of LSTM cells, using the output of the encoder and the last stroke it produces to generate the next stroke. The initial stroke is fixed as (0, 0, 1, 0, 0), indicating that it is a base point and the current stroke should be continued. The output format of the decoder is 
\begin{math}
(\omega, \mu_x, \mu_y, \sigma_x, \sigma_y, corr_{xy}, \rho)
\end{math}. 
The Gaussian mixture model is involved to generate the next point of the stroke by sampling.

For the training part, Sketch-RNN uses two loss functions. The first is Reconstruction Loss L$_{R}$, and the other is Kull-Leibler Divergence Loss L$_{KL}$ According to \cite{c_3}, The Loss function can be written as:
\begin{equation}
   L(S) = E_A (z|S) [\log B(S' | z)] - D_{KL} (A(z|S) || B(z)) 
\label{eq_1}
\end{equation}

\noindent
The input S is the sequence of strokes. The first part E$_{A}$(·) is the Reconstruction Loss L$_{R}$, which measures the similarity between the generated sketches and the original sketches in the training set. The second part D$_{KL}$(·) is the Kull-Leibler Divergence Loss L$_{KL}$, which compares the distribution of the generated strokes and those in training set. A(·) and B(·) correspondingly represent the output of the encoder and the decoder.

\begin{figure}[ht]
\centering
\includegraphics[width=0.8\textwidth]{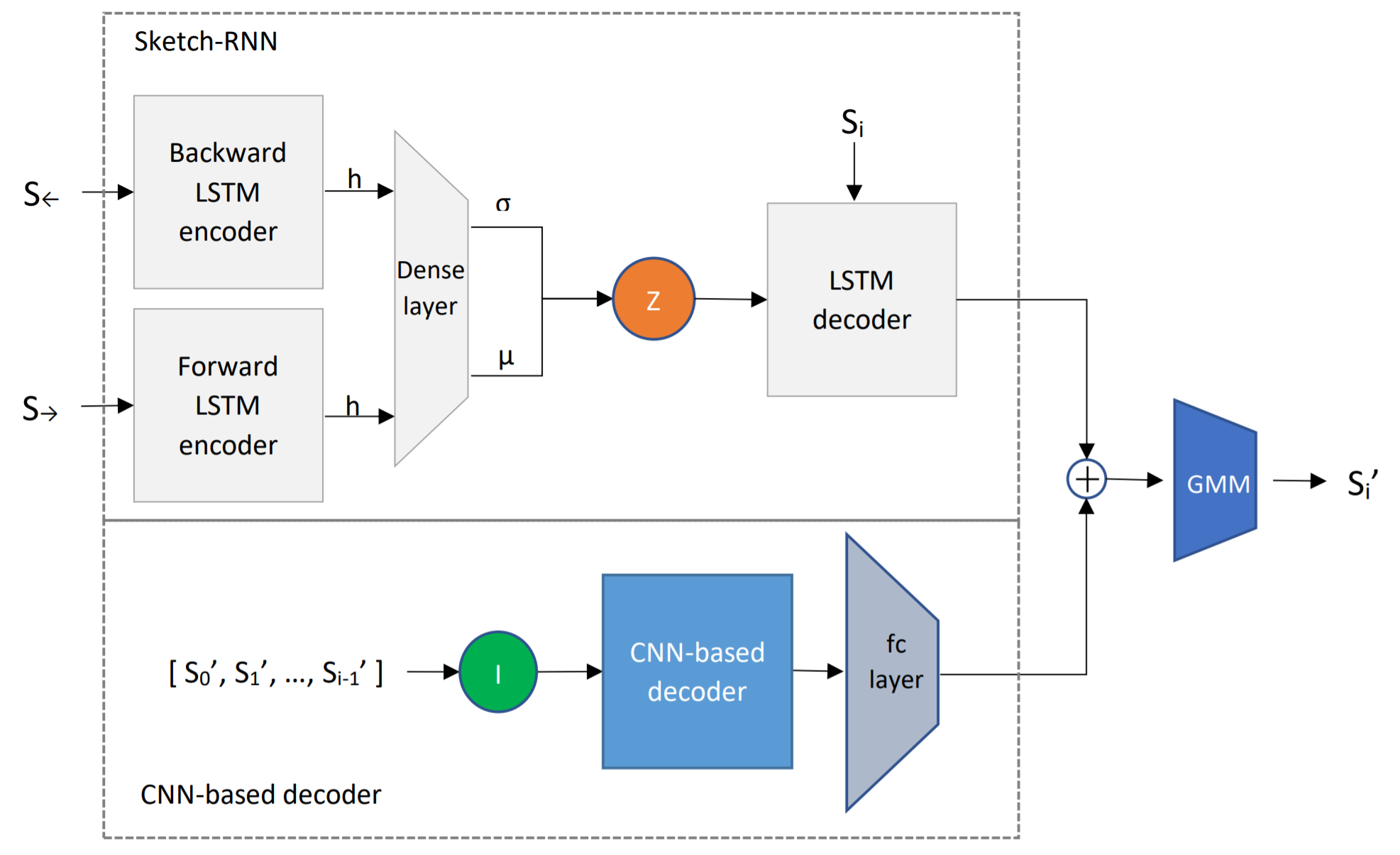}
\caption{VAE model we proposed}
\label{f_1}
\end{figure}

\subsubsection{Our Model}
\noindent
As is shown in Fig.~\ref{f_1}, our model extends the original model. To enhance the quality of sketches produced by Sketch-RNN, we added an extra CNN-based decoder. This CNN-based decoder is parallel to the original decoder. Instead of producing any new strokes, the CNN decoder only adjusts the position of the strokes generated by Sketch-RNN. We first applied weight on both the CNN decoder output and the RNN output. Then we added them together and used a Gaussian mixture model to sample the predicted next point of the stroke.

\begin{figure}[ht]
\centering
\includegraphics[width=0.8\textwidth]{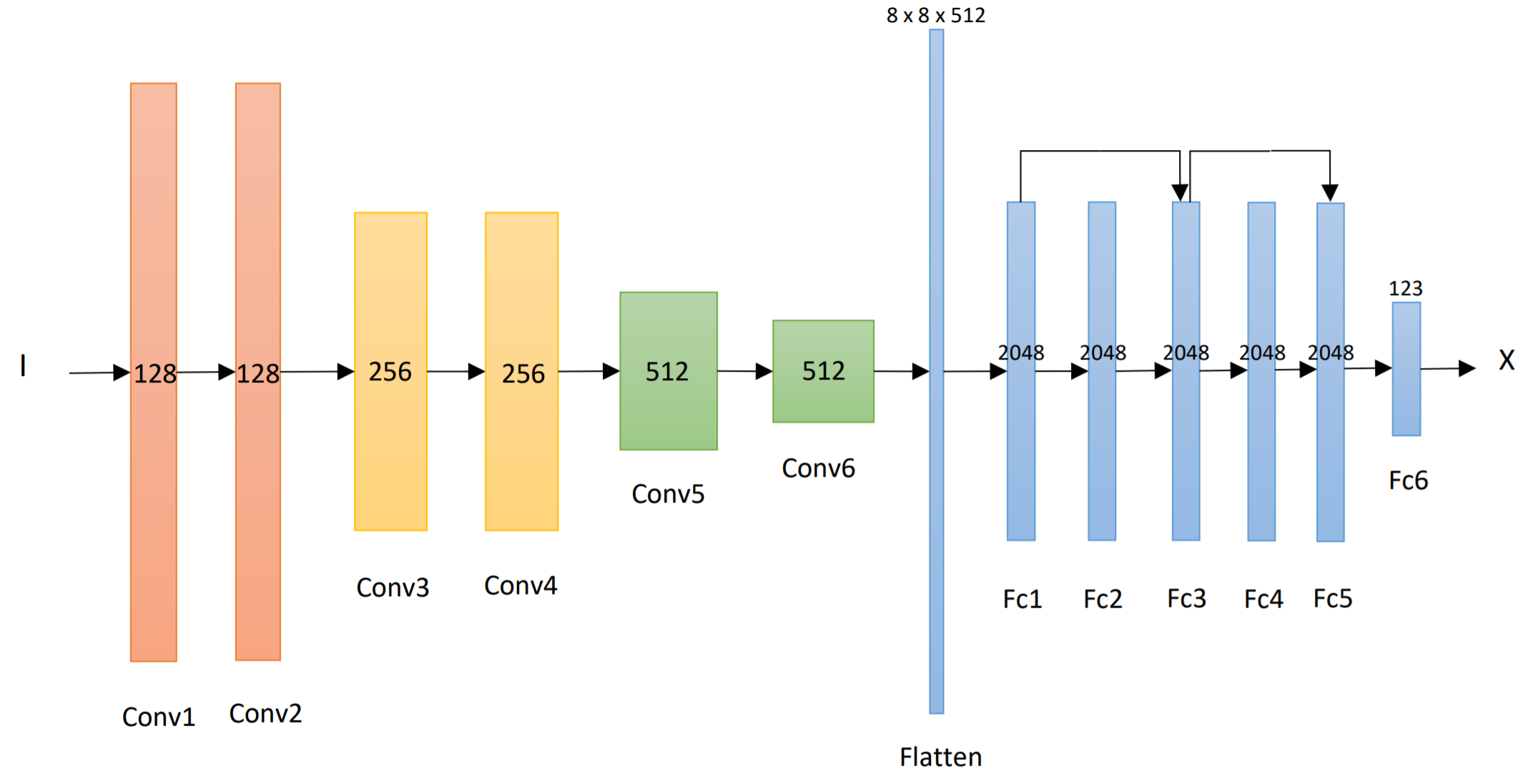}
\caption{CNN-based decoder}
\label{f_2}
\end{figure}

The structure of the CNN model is shown in Fig.~\ref{f_2}, with 6 convolutional layers, 1 flatten layer, and 3 dense layers. The input is an image, which gathers all previous generated strokes [S$_{0}$, S$_{1}$, …, S$_{t-1}$] and transforms them into a sketch-like image. The size of the image is [128, 128], which is a trade-off between the efficiency of computing and the clearness of the features in the image. The output is a sequence in the format of
\begin{math}
(\omega, \mu_1, \mu_2, \sigma_1, \sigma_2, corr_{xy}, \rho)
\end{math}
, which is the same as that of the original decoder.

For the detail of the CNN implementation, we adjusted our model by trial and error. The depth for each kernel on each convolutional layer is (3, 128, 128, 256, 256, 512), with strides (1, 2, 1, 2, 2, 2) on each layer. We then flattened the last output of the convolutional layer and used the structure of residual network \cite{c_11} to extract features in the image and generate the latent distribution for the next stroke point. In the convolutional layers we used ReLU (Rectified Linear Unit) \cite{c_17} as the activation function. ELU (Exponential Linear Unit) \cite{c_16} was applied in dense connected layers.

\subsection{Loss Function}
\noindent
The loss function is the sum of two separated parts. The first part is the point offset Loss L$_{S}$. This loss function minimizes the offset of the predicted point 
\begin{math}
(\Delta_x,\Delta_y)
\end{math}
:
\begin{equation}
\begin{aligned}
    L_S = -\frac{1}{S_{max}}\sum_{i=1}^{S_{stop}} log (\sum_{j=1}^M \prod_{j,i} \mathcal N (\Delta_{x_i}, \Delta_{y_i} | \mu_{x,j,i}, \mu_{y,j,i}, \sigma_{x,j,i}, \sigma_{y,j,i}, Corr_{xy,j,i}))
\label{eq_2}
\end{aligned}
\end{equation}
\noindent
The generated offset 
\begin{math}
(\Delta_x, \Delta_y)
\end{math}
is discarded if the length of the generated sequence is longer than S$_{stop}$. S$_{stop}$ is controlled by the output (p$_{1}$, p$_{2}$, p$_{3}$) of the decoder where
\begin{math}
\sum_{i=1}^3 p_i = 1
\end{math}
. The other loss function L$_{P}$ minimizes the difference between the generated pen states (p$_{1}$, p$_{2}$, p$_{3}$) and these of labels:
\begin{equation}
    L_P = -\frac{1}{S_{max}}  \sum_{i=1}^{S_{max}} \sum_{k=1}^3 P_{k,i} log(q_{k,i})
\label{eq_3}
\end{equation}
Unlike L$_{S}$, L$_{P}$ is calculated through all the output (p$_{1}$, p$_{2}$, p$_{3}$) of the decoder until it reaches the max sequence length.q$_{k, i}$here means the categorical distribution of the ground truth (p$_{1}$, p$_{2}$, p$_{3}$), where 
\begin{math}
\sum_{k=1}^3\sum_{i=1}^{S_{max}} q_{k,i} = 1
\end{math}
. The total loss function is:
\begin{equation}
    L = L_S + L_P
\label{eq_4}
\end{equation}
This loss function is applied to train CNN-based decoder. To ensure stability and quality of generated strokes, we use Sketch-RNN model pretrained on cat sketch data set.

\subsection{Training}
\noindent
We used cat sketches to train the CNN model. The data set is divided into 10,000 training samples, 2,500 test samples and 2,500 validation samples. For each sketch, we randomly cropped the sequence of strokes into two parts. We used the first part as the input to the CNN-based decoder and the second part as labels.

Normalization was used to pre-process the images. The normalization process is defined as:
\begin{equation}
    (r^\prime, g^\prime, b^\prime) = 1 - (r, g, b) / 255.0
\label{eq_5}
\end{equation}
\noindent
By normalization, dark strokes are mapped to 1's while background is mapped to 0's, which is easier for convolutional layers to extract features from the images.

\begin{figure}[htbp]
\centering
\includegraphics[width=\textwidth]{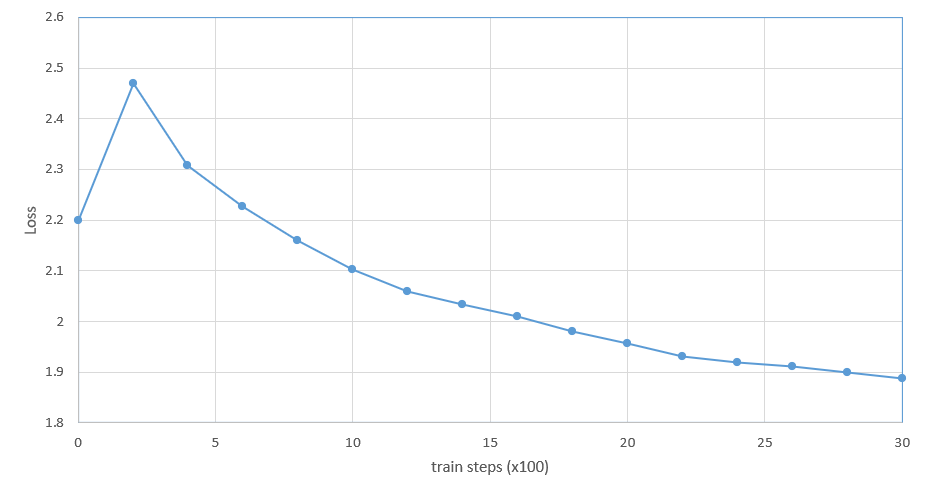}
\caption{Loss value on train steps}
\label{f_3}
\end{figure}

To ensure the training process is stable, we have adjusted the learning rate to \begin{math}
10^{-6}
\end{math}. Loss value fluctuation is shown in Fig.~\ref{f_3} as train iterations.

\begin{figure}[htbp]
\centering
\includegraphics[width=\textwidth]{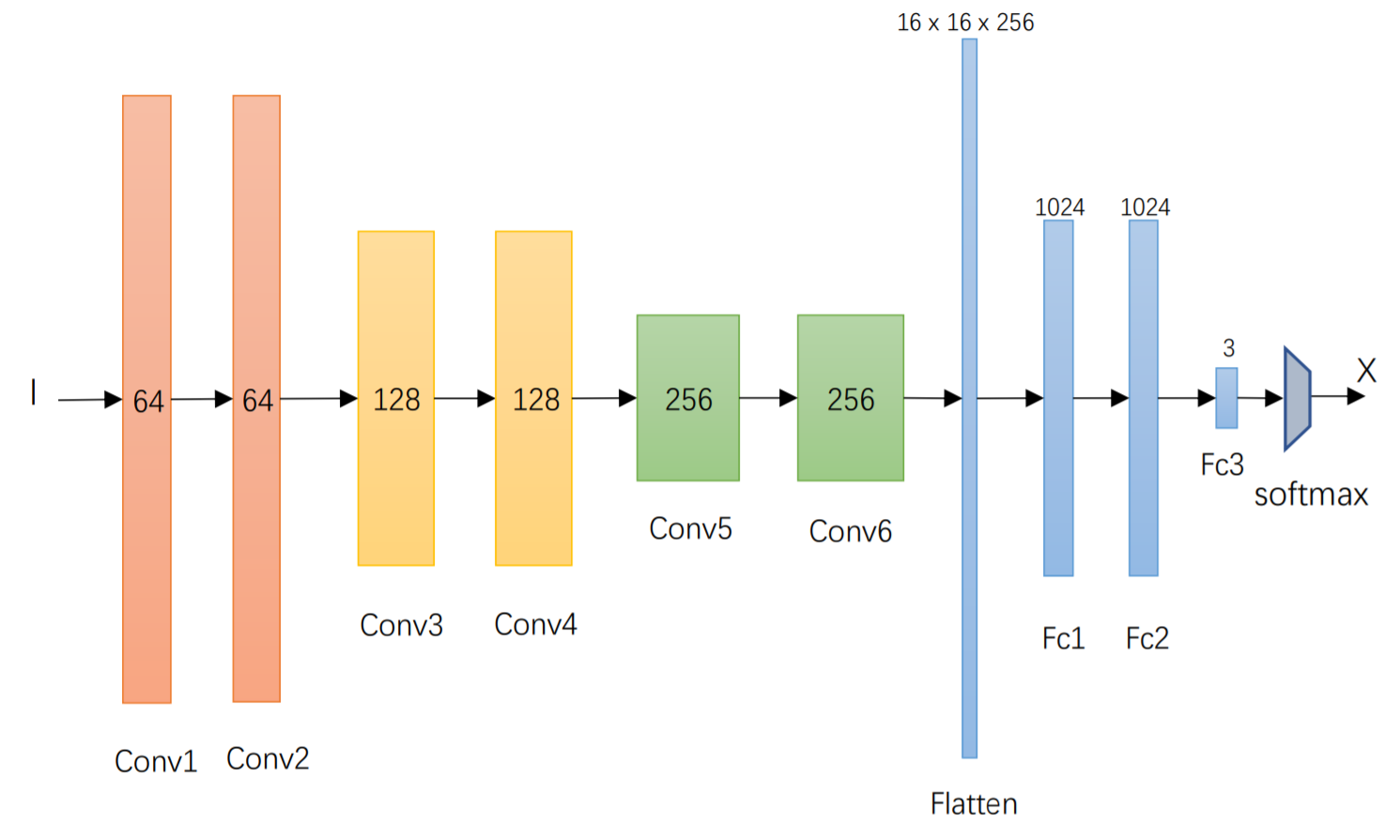}
\caption{CNN-based Discriminator}
\label{f_4}
\end{figure}

\subsection{Evaluation}
\noindent
To examine whether our model could make better adjustments to the generated sketch, we designed two different approaches to evaluate the performance of our model.

\subsubsection{Discriminator Model}
\noindent
Inspired by the Discriminator in GAN \cite{c_7}, we used a CNN-based model to judge whether a generated sketch is closer to human-drawn sketches or not. The goal for this CNN-based model is to accurately distinguish among three classes: sketches from Sketch-RNN, sketches from our model, and human-drawn sketches. Our model succeeds when it misleads the CNN-based model and the sketches generated by it are classified as human-drawn sketches. Additionally, if the discriminator fails to precisely discern sketches generated between Sketch-RNN and our model, it indicates that the features of sketches produced by our model have no significant difference compared to those produced by Sketch-RNN.

As is shown in Fig.~\ref{f_4}, the input I denotes image input to the model. The output X the output, is a three-dimensional vector, representing three categories: sketches generated by Sketch-RNN, sketches generated by our model, and sketches drawn by humans. The model uses six layers, with (64, 64, 128, 128, 256, 256) kernels for each layer. The size of kernels on each layer is 
\begin{math}
[3 \times 3]
\end{math}
. The stride for each layer is (1, 2, 1, 2, 1, 2). Activation function used in convolution layers is ReLU while that used in fully connected layers is ELU.

We trained the Discriminator Model on a total of 30,000 cat sketches, including 10,000 sketches generated by Sketch-RNN, 10,000 produced by our model, and 10,000 drawn by human. The validation set contains 3,000 cat sketches, including 1,000 Sketch-RNN generated sketches, 1,000 our model generated sketches, and 1,000 human-drawn sketches. We achieved 95.2\% accuracy on recognizing human-drawn sketches, 70.3\% accuracy on recognizing sketches from Sketch-RNN, 63.2\% accuracy on recognizing sketches from our model. The result analysis will be discussed in section 4.2.1.

\subsubsection{t-SNE}
\noindent
T-distributed Stochastic Neighbor Embedding (t-SNE) \cite{c_9} is applied to measure the similarity of features in the sketches. This method maps a high-dimensional data set into a low-dimensional space, which is convenient for visualization. Similar objects will be placed close to each other while dissimilar objects will be distantly located.

We used t-SNE to analyze our model to answer two questions (compared with Sketch-RNN):

\begin{enumerate}
[topsep=0pt,itemsep=1pt,partopsep=0pt,label=\arabic*)]
\item Whether our model exerted significant change to the distribution of strokes;and
\item Whether our model reduces disordered strokes. Or, which model is able to create sketches with more creativity.
\end{enumerate}

In our experiment, We randomly sampled 1,000 cat sketches from Sketch-RNN and 1,000 cat sketches from our model. We plotted a scatter graph to visualize the result of dimension reduction.

\section{Experimental Results}
\subsection{Image Results}
\noindent
When comparing images generated by our model with those produced by Sketch-RNN, we found out an interesting feature in our model: it tended to add 'eyes' on the face of cat sketches. Figure 5 displays some images that illustrate this character, where the images labeled 'Sketch-RNN' were produced by Sketch-RNN, while images labeled 'Proposed' were produced by our model.

\begin{figure}[htbp]
\centering
\includegraphics[width=0.8\textwidth]{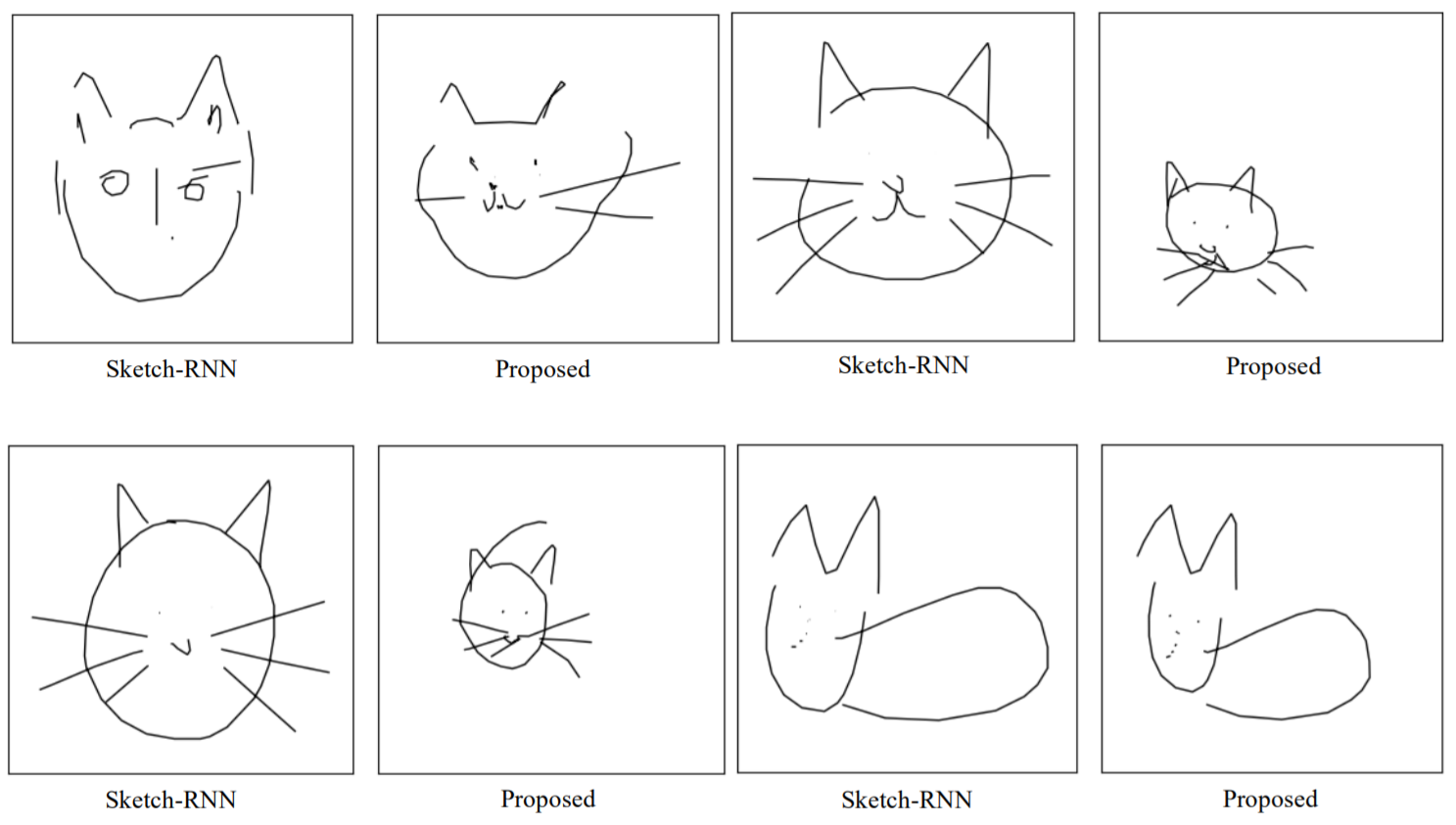}
\caption{Comparison of sketches generated by Sketch-RNN and our model}
\label{f_5}
\end{figure}

\begin{figure}[htbp]
\centering
\subfigure[Sketch-RNN]{               %小图题的名称
\includegraphics[width=0.8\textwidth]{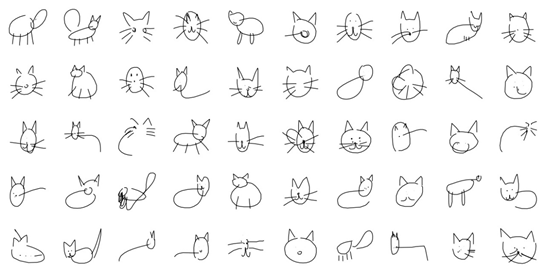}
\label{f_6a}}
\hspace{0in}
\subfigure[Proposed]{
\includegraphics[width=0.8\textwidth]{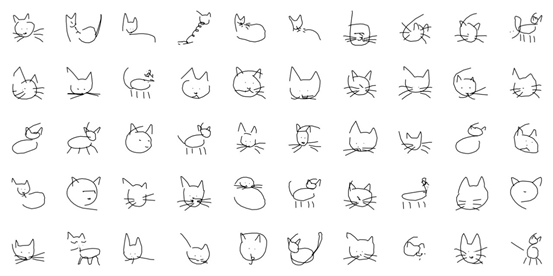}
\label{f_6b}}
\caption{Sketches generated by Sketch-RNN and our model}
\end{figure}

The cartoon-like eyes generated by our model indicated that our model has the ability to understand what the network is drawing. More importantly, our model can recognize what features the original sketches have while Sketch-RNN forgets to present.

To further illustrate the quality of sketches our model generated, we present 50 sketches generated by each model randomly picked from a total of 2,000 sketches. Sketches generated by Sketch-RNN (Fig.~\ref{f_6a}) and our model (Fig.~\ref{f_6b}) are shown.

\subsection{Evaluation Results}
\subsubsection{Discriminator}
\noindent
Firstly, we used the Discriminator as described in Section 3.5.1. We randomly sampled 1,000 Sketch-RNN generated sketches, 1,000 our model generated sketches, and 1,000 human-drawn sketches as validation set. Table 1 shows the percentage of classification results by Discriminator. The first column represents the actual class sketches belong to. The first row denotes the predicted class of sketches made by the Discriminator.

\renewcommand{\arraystretch}{1.3}
\begin{table}[htbp]
\centering
\caption{Classification results by Discriminator}
\begin{tabular}{|l|m{0.22\textwidth}|b{0.22\textwidth}|p{0.22\textwidth}|}
\hline
\diagbox{\rule{2pt}{0pt} Label \rule{12pt}{0pt}} {\rule{12pt}{0pt} Predict  \rule{2pt}{0pt}} & \rule{2pt}{0pt} Sketch-RNN & \rule{2pt}{0pt} Our model & \rule{2pt}{0pt} Human-drawn \\[5pt]
\hline
\rule{2pt}{0pt} Sketch-RNN \rule{0pt}{15pt} & \rule{2pt}{0pt} 70.6\% & \rule{2pt}{0pt} 22.0\% & \rule{2pt}{0pt} 7.4\% \\[5pt]
\hline
\rule{2pt}{0pt} Our model \rule{0pt}{15pt} &	\rule{2pt}{0pt} 23.2\% & \rule{2pt}{0pt} 63.2\% & \rule{2pt}{0pt} 13.6\% \\[5pt]
\hline
\rule{2pt}{0pt} Human-drawn \rule{0pt}{15pt} & \rule{2pt}{0pt} 0.9\% & \rule{2pt}{0pt} 3.9\% & \rule{2pt}{0pt} 95.2\% \\[5pt]
\hline
\end{tabular}
\label{t_1}
\end{table}

Classification accuracy for each of the three classes: ‘Sketch-RNN’, ‘Our model’ and ‘Human-drawn’ sketches is 70.6\%, 63.2\% and 95.2\% correspondingly. The model could not distinguish between sketches from Sketch-RNN and our model very well. This indicates that the features of sketches generated by our model are similar to those generated by Sketch-RNN.

Table~\ref{t_1} also illustrates the ability of a model to mislead the discriminator to view the sketch as the original one drawn by humans. Our model has a 13.6\% of probability to successfully mislead the discriminator. Sketch-RNN has a 7.4\% of probability to confuse the discriminator. Therefore, our model produces sketches with more features that the original sketches possess.

\subsubsection{t-SNE}

\begin{figure}[ht]
\centering
\includegraphics[width=0.8\textwidth]{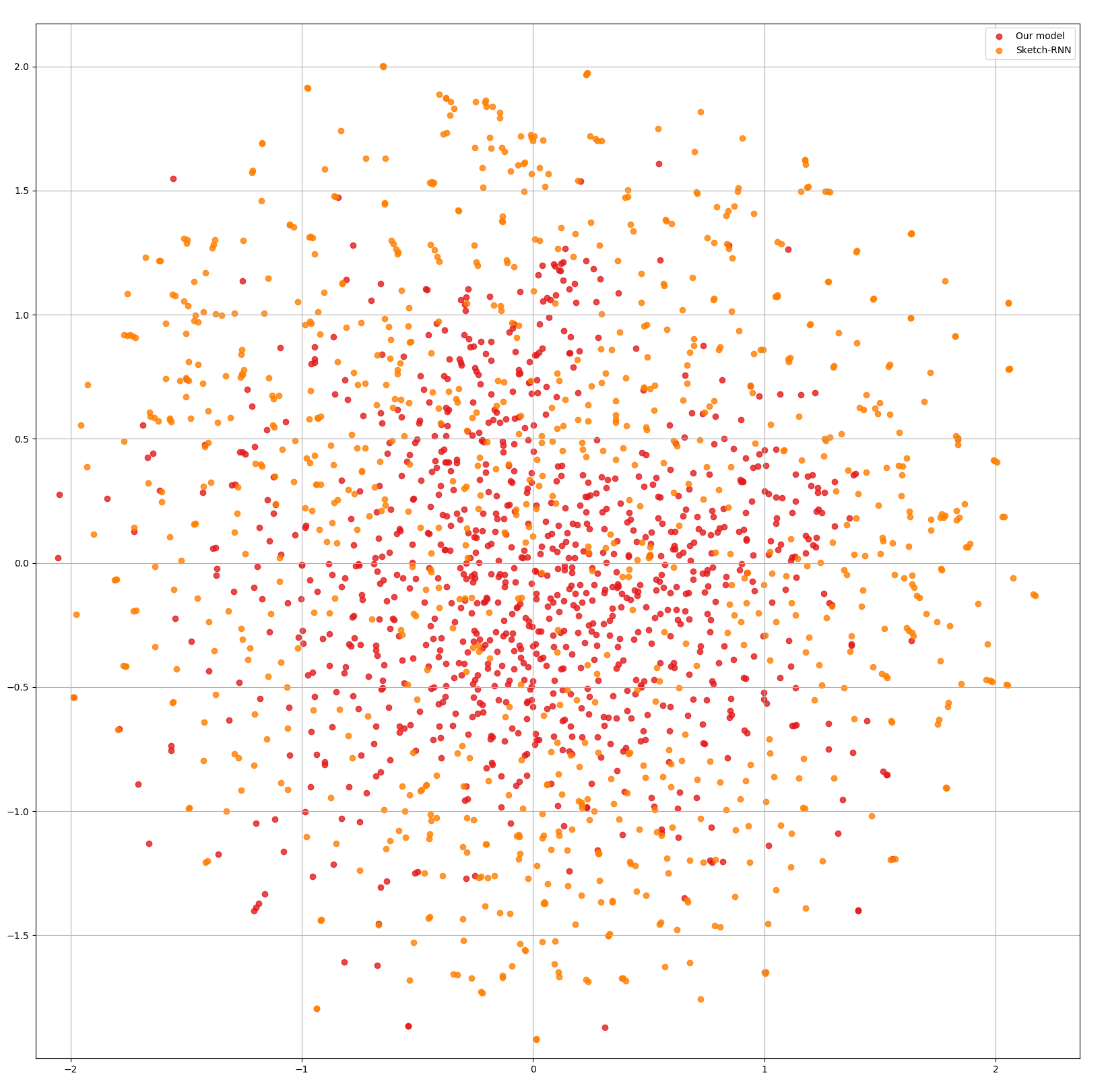}
\caption{t-SNE result}
\label{f_7}
\end{figure}

\noindent
As is shown in Fig.~\ref{f_7}, the red dots are sketches generated by our model. The orange dots are sketches generated by Sketch-RNN. The scatter graph shows that sketches produced by our model are more concentrated than those produced by Sketch-RNN. This indicates that the features in the cat sketches from our model is more unified. Hence our model generates sketches with more stable features while Sketch-RNN is more creative, which is able to generate sketches with more feature diversity.

Compared with the results in section 4.2.1, Our model has learned some features from human-drawn sketches. Therefore, our model is able to confuse the discriminator to judge its sketches as human-drawn with higher probability. Our t-SNE analysis suggests that our model applies the features it learned from human-drawn sketches more often. Combining these two results together, we can conclude that our model reduces the number of disordered strokes and produces sketches that are more similar to human-drawn sketches.

\section{Conclusion}
\noindent
In this paper, we introduce a CNN-based decoder, a model that improves the quality of sketches drawn by Sketch-RNN. The decoder learns from incomplete human-drawn sketches and predicts the next stroke. Both the Discriminator model and t-SNE were used for evaluation. We compared the evaluation results between our model and Sketch-RNN, and found that our model produces sketches with higher quality. However, since the CNN-based decoder has to produce predictions on each iteration of generation, our model is four times slower than Sketch-RNN. We have only tested our model on a data set of cat sketches. More experiments on different sketch data sets need to be conducted.

\section{Acknowledgement}
\noindent
We would like to express our great appreciation to Professor Gregory Kesden, Carnegie Mellon University, for his constructive suggestions and patient guidance. We would also like to thank Kexin Feng, Ph.D. student at Texas A\&M University, and Naijing Zhang, student at UC Berkeley, for their encouragement and critiques for this project.

\bibliographystyle{splncs04}
\bibliography{references}

\end{document}